# LANGUAGE ACCESS: AN INFORMATION BASED APPROACH

Akshar Bharati
Vineet Chaitanya
Amba P. Kulkarni
Rajeev Sangal
Language Technologies Research Centre
Indian Institute of Information Technology Hyderabad
**{vineet,amba,sangal}@iiit.net**

## ABSTRACT

The anusaaraka system (a kind of machine translation system ) makes text in one Indian language accessible through another Indian language. The machine presents an image of the source text in a language close to the target language. In the image, some constructions of the source language (which do not have equivalents in the target language) spill over to the output. Some special notation is also devised.

Anusaarakas have been built from five pairs of languages: Telugu,Kannada, Marathi, Bengali and Punjabi to Hindi. They are available for use through Email servers.

Anusaarkas follows the principle of substitutibility and reversibility of strings produced. This implies preservation of information while going from a source language to a target language.

For narrow subject areas, specialized modules can be built by putting subject domain knowledge into the system, which produce good quality grammatical output. However, it should be remembered, that such modules will work only in narrow areas, and will sometimes go wrong. In such a situation, anusaaraka output will still remain useful.

## 1. INTRODUCTION

Fully-automatic general purpose high quality machine translation systems (FGH-MT) are extremely difficult to build. In fact, there is no system in the world for any pair of languages which qualifies to be called FGH-MT. The reasons are not far to seek. Translation is a creative process which involves interpretation of the given text by the translator. Translation would also vary depending on the audience and the purpose for which it is meant. Since at present, the machine is not capable of interpreting a general text with sufficient accuracy automatically - let alone re-expressing it for a given audience, it fails to perform as FGH-MT. The main difficulty that the machine faces, pertains to dealing with ambiguity. A given text *codes* only a part of the *information*. Ambiguity is resolved (by guessing) using world knowledge, domain specific knowledge, etc., a task which turns out to be very difficult for the machine.

## 1.1 INFORMATION CODING

To understand the idea that the text expresses only a part of the information, let us consider an example. In Indian languages, which have relatively free word-order, information that relates an action (verb) to its participants (nouns) is primarily expressed by means of post-positions or case endings of nouns (collectively called vibhaktis of the noun). For example, in the following sentence in Hindi:

rAma ne roTI khAI (1)
Ram erg. bread ate
Ram ate the bread.

The ergative (erg.) post position marker ('ne') after 'rAma' indicates that Ram is the *karta* of eat, which here means that Ram is the *agent* of eating. (Note that in English, the primary device for expressing the same information is by means of word order.)

Noun-verb agreement also helps in identifying the karta.
For example, in the following sentence:

rAma roTI khAtA HE (2)
Ram(m.) bread(f.) eats(m.)
Ram eats bread.

the masculine (m.) ending of the verb indicates that the karta is masculine, which in this sentence unambiguously means Ram. However, this is not always unambiguous; consider the following sentence:

chAvala rAma khAtA hE (3)
rice (m.) Ram (m.) eats (m.)
Ram eats rice.

in which the agreement does not help in identifying the karta unambiguously, because there are two masculine nouns (Ram and rice) one of which is the karta. Translation to English, say, would be quite different depending on which one is the karta.

In language, there is a tension between brevity and ambiguity. If everything was explicity stated, the text would be less ambiguous but would be long. Brevity also helps in focussing attention to the relevant parts. Ambiguity seems to be a necessary price for conciseness and focus.

## 1.2 FAITHFULNESS vs. NATURALNESS

To build a practical MT system, the load has to be shared between man and machine. A clean way to share the load is for the machine to take up the task of language related processing, and to leave the processing related to background knowledge to the reader. Language related processing consists of analysis of the input source language text such as morphological processing, use of bilingual dictionary, and any other language related analysis or generation. These are the primary sources of difficulty to the reader. These are also the tasks which are relatively easier for the machine. On the other hand, world knowledge related aspects are left to the reader, who is naturally adapt at it.

In translation, two opposing forces are at work: faithfulness and naturalness. The translator must chose between faithfulness to the original text and naturalness to the reader. Most translations that we come across, are weighted towards naturalness to the reader. Anusaaraka is at the other extreme: it tries to be as faithful to the original text as possible. In fact, its output must contain all the information in the source language text, and should have no other new information.

There is a problem in coding "exactly" the same information (with 100% fidelity) from one language to another, particularly if we want to generate sentences of about equal length, paralleling the sentence constructions wherever possible. (In this sense, translation is sometimes said to be an impossible task). FOOTNOTE{This also suggests the incommensurability of information, discussed later in this paper.}

## 1.3 ANUSAARAKA or LANGUAGE ACCESS

The anusaaraka answer lies in deviating from the target language in a systematic manner whenever necessary. This new language is something like a dialect of the target language. The anusaaraka output can be said to be the image of the source text, much like what the camera produces. Reading the image of source text is like reading the original text. It will have the same flavour. Translation, on the other hand, is like a painting. The translator interprets the original in the source language, and "paints" a text in the target language with approximately the same meaning and nuance.

Readers will usually require some learning of the dialect of the target language (discussed in the next section). This learning time will be negligible compared to the learning time of the source language.

Indian languages are relatively free word-order where the noun-groups can come in any order followed generally by the verb group. (The order conveys emphasis etc. but not the information about karaka relationships or theta roles.) If we take a sentence in a source language, and substitute the word groups in it by appropriate word groups in the target language, it works well because the languages make similar use of order to convey emphasis etc. The vibhaktis for the word groups (that is, case endings and post-position markers for nouns, and TAM for the verb groups), must be mapped from the source language to the target language carefully, as they contain important karaka information regarding the verb and the nouns. Again the languages behave in a similar way.

Besides the above, there are similarities in the meanings of words. Many words in the languages have a shared origin (from Sanskrit), and because of shared culture, they usually also share meanings. This implies that for a source language word, the bilingual dictionary provides a unique answer in the target language for a large percentage of words (80% for Kannada to Hindi).

Now, we will discuss some problems because the two languages differ, and see how these problems can be handled. We will take examples from Hindi, Telugu and Kannada, three of the major languages of India. Hindi is spoken by more than half a billion people, Telugu by about 80 million, and Kannada by about 50 million people.

Telugu and Kannada differ from Hindi, as they are Dravidian languages, and are further apart from Hindi compared to other Indian languages. Even then, except from agreement, there are only three major syntactic differences between Hindi and Kannada. Surprisingly all of these can be taken care of by enriching Hindi with a few additional functional particles or suffixes as shown below. Thus, they can be viewed as lexical gaps or function word gaps.

## 2. ANUSAARAKA - LANGUAGE BRIDGES

We now take up some important constructions in the south Indian languages, which differ from Hindi, and show how they have been bridged in anusaaraka.

### 2.1 COMP ("ki") CONSTRUCTION

In case of embedded sentences in Hindi, the subordinate sentence is put after the main verb unlike in Kannada. For example (where, label H is for Hindi, !E is for English gloss):

```
  H: rAma ne  kahA ki   mEM  ghara ko jAUMgA.            (4)
 !E: Ram erg. said that I    home acc. will_go
  E: (Ram said that he will go home.)
```

There is a construction in Kannada using 'eneMdare' or 'that' which is similar, but is seldom used. Kannada uses another construction for which the anusaaraka Hindi is given below.

```
  K: mohana nALe     baruvanu  eMdu  rAma heLidanu.   (5)
 @H: mohana kala     AyegA    EsA   rAma kahA.
 !E: Mohana tomorrow come-fut this  Rama said.
```

Although, 'EsA' construction is a proper construction in Hindi; it is seldom used. In the dialect of Hindi produced by anusaraka from southIndian languages however, this will be the normal construction used.

## 2.2 RELATIVE-CLAUSE ("jo") CONSTRUCTION

In this section, we will discuss how anusaaraka handles participle verbs (behaving as adjectives) in Telugu to produce the same information in Hindi. The solution works for all south Indian languages, which display this phenomenon.

We will first try to arrive at the information contained in TAM labels which stand for adjectival participle, in a mathematically precise way. Let us take the following Telugu example sentence:

```
  T: rAmuDu tinina camacA veVMDidi.                    (6)
     ------ --- -- ------ --------
       1    2a  2b    3      4
 !E: Ram   *eaten  spoon  of-silver
  E: The spoon with which Ram ate is of silver.
```

(* 'eaten' is only an approximation, 'tinina' is a past-participle form of 'tina' or 'eat')

We are interested in finding the meaning of the TAM label or suffix 'ina' suffix in 'tinina' above. Let us name it 2b, and the rest of the words are also named for easy reference.

If a Telugu-Hindi bilingual person is asked to translate the sentence, he is likely to write down the following in Hindi:

```
  H: rAma ne  jisa  cammaca se   khAyA, vaHa cAMdI kA HE.
     ---- ++        -------      ---         -------- ++
      1             3            2a           4
 !E: Ram erg. which spoon instr. ate,   that silver_of is
  E: The spoon with which Ram ate is of silver.
```

Here the Hindi words are marked corresponding to the Telugu words (other than 2b whose value we want to find out). '++' is used to denote words that have been put by the translator but which are not there in the original Telugu sentence. 'ne' corresponds to the ergative marker which is an idiosyncracy of Hindi. Also it is known that 'HE' at the end (copula) is mandatory in the Hindi sentence but is absent in the given Telugu sentence.

We can rephrase the sentence in Hindi to get the words in the same order:

```
  H: rAma ne jisa se khAyA HE vaHa cammaca cAMdI kA HE.
     ---- ++         ---         ------- -------- ++
      1              2a            3        4
```

or better still, we may rewrite the above as:

```
 H: rAma ne khAyA  HE  jisa  se     vaHa cammaca cAMdI kA HE.   (7)
    ---- ++ ---                         ------- -------- ++
      1     2a                            3       4
!E: Ram erg. eaten has which instr. that spoon   silver_of is
```

wherein the order of the words including the parts of words (2a and 2b) is exactly the same as the order in the original sentence. Now the part which remains unassigned, stands for 2b. Therefore, we get the equation:

```
 ina = yA_HE_jisa_se_vaHa
      -en_is_which_instr_that (English gloss)
       has_VERB_en_with_which_that (English explanation)
```

But a closer scrutiny reveals an assumption, "se" or instrumental marker is not there in the Telugu sentence. For example, consider the following sentence:

```
 T: rAmuDu winina pleTu  veVMDidi                              (8)
    ------ --- -- ------ --------
      1    2a  2b    3      4
!E: Ram    eaten  plate  silver-of
 E: The plate in which Ram ate is of silver.
```

Its equivalent Hindi sentence is:

```
 H: rAma ne khAyA HE jisa meM vaHa pleTa cAMdI kI HE.     (9)
    ---- ++ ---                  ----- -------- ++
      1     2a                     3      4
```

The above sentence yields the following equality:

```
 ina = yA_HE_jisa_meM_vaHa
      -en_is_which_loc_that (English gloss)
       has_VERB_en_in_which_that (English explanation)
```

The two different equalities for 'ina', and similar other examples lead us to conclude that the 'se' or 'meM' markers are not there in the 'ina' but are supplied by the reader based on the world knowledge. Therefore, the equality becomes:

```
 ina = yA_HE_jo_*_vaHa
```

where '*' stands for an unspecified post-position to be supplied based on context. After further refinement (not discussed here), it becomes:

ina = yA_[HE/tHA]_jo_*_vaHa-
-en_[is/was]_which_*_that (English gloss)

The claim is that the above is a mathematically precise equivalence between the 'ina' Telugu TAM and anusaaraka Hindi.

The above can be restated as follows: It shows the equivalence between the adjectival participle in Telugu and the relative clause in Hindi, which has been known, but which the above equation makes precise. Although, Hindi also has participial phrases, it has only two TAMS: yA and tA_HuA (with perfective and continuous aspects, respectively).

H: khAyA HuA phala (10)
eaten fruit

H: khAtA HuA hiraNa (11)
eating deer

As a result, these are not sufficient to capture other TAMs which might occur in Telugu. (There are syntactic holes in participles in Hindi.)

There is another problem, too, as we have seen. The two participial phrases in Hindi have coding for karaka relations (theta roles) which is absent in Telugu. TAM 'tA_HuA' codes karta karaka (roughly agent), and the sentence (11) says, the deer who is eating (not the one who is being eaten). Similarly, 'yA' codes karma as in sentence (10) (the fruit being eaten, and not the fruit who is eating). FOOTNOTE{More correctly, 'yA' codes karma in case of sakarmaka or transitive verbs, and karta in case of intransitive verbs.} Thus, Hindi is poorer than Telugu in coding tense, aspect, modality information, while richer in coding karaka information. But this creates another difficulty for anusaaraka. Using these constructions in Hindi, would mean putting in something that is not contained in the source language sentence, and the information equivalence would be lost.

Unlike the 'ki' construction (Section 2.1), this idea takes some time and effort for the Hindi reader to get used to.

Another construction, not discussed here for want of space, is the "ne" construction or ergative marker, which is a peculiarity of only the Western belt languages in India. Therefore, while building the anusaaraka from south Indian language to Hindi, such a construction would not occur in the output.

## 2.3 PRE-EDITING AND POST-EDITING

Anusaaraka system has been designed so that the combination of man and machine together can perform translations, etc. The user can help in pre-editing the input and post-editing the output. In the pre-editing task, the input text is corrected and edited by the user: Words spelt with non-standarad spellings are changed to their standard spellings, external sandhi between words is broken (unless it changes meaning), etc. This is an important task for Indian Languages because of lack of standardization and consequent variation.

Similarly, post-editing can be carried out on the output produced by the machine. There are three levels of post-editing. The first level of post-editing seeks to make the output grammatically correct. The emphasis is on speed and low cost.

In the second level of post-editing the raw output is corrected not only grammatically but also stylistically. For example, 'Esa' construction would be changed to 'ki' (see Section 2.1). In the third level of post-editing the post-editor might change the setting and the events in the story to convey the same meaning to the reader who has a different cultural and social milieu. This is really trans-creation, and a creative post-editor (who can even be mono-lingual) can go all the way upto this level.

## 3. ANUSAARAKA PROCESSING

Anusaaraka processing could also be viewed as a series of *information preserving* transformation, to bring the source language close to the target langauge.

Information preserving transformations follow two properties:
  A. substitutivity, and
  B. reversibility.

### 3.1 SUBSTITUTIVITY

This basically takes care of one-to-many mapping. When a word or a phrase has two equivalent meanings, both alternatives are put in the output, unless one is ruled out by local word grouping.

For example, Hindi word 'khAtA' can be replaced with two possible English words:
   khAtA -> eats/ledger

If Hindi morphological analyser replaces 'khAtA' with 'eats' because it is the more frequent usage, then substitutivity will be violated in the following sentence:

```
  H: rAma ne  bEnka meM apanA khAtA   kholA.          (12)
 !E: Ram erg. bank  in  his   ACCOUNT opened.
  E: Ram opened his account in the bank.
```

Basic idea is that all possible substitutions must be exhaustively enumerated. (Original ambiguities must be carried over.)

Substitution rule can use context, but the rule should be universal, i.e., should work in all possible contexts. (No guessing.)

Non-trivial example: Participles in south Indian languages, have been already discussed in detail in Section 2.2:

  _ina -> _yA_[hE/thA]_jo_*_vaha- (Telugu to Hindi)

### 3.2 REVERSIBILITY

The transformation should also be reversible. It should be possible to go back from transformed string to initial string.

This takes care of many-to-one mapping, and the basic idea behind this principle is that the information should not be thrown away. We illustrate it by an example from Telugu to Hindi:

```
  A/adi/vADu/AmeV -> vaHa
                     he/she/that (English)
```

Single 'vaHa' for all four might seem natural and appealing. In fact, most MT systems in such a setting would be happy that they did not have to chose out of alternatives, and would simply use 'vaHa'. However, throwing away information is NOT good. For example, the sequence 'vaHa ghara' might stand for a single phrase, or two noun phrases:

```
  vaHa ghara  --> that house

   H: vaHa ghara acchA HE.          (13)
  !E: That house good  is
   E: (That house is good.)

  vaHa ghara gayA --> He went home.

   H: vaHa ghara gayA.              (14)
  !E: He   home went.
   E: (He went home.)
```

While in the above examples, the difference in meaning is readily apparent, it might not always be so. On the other hand, there is a different source word in Telugu for the two cases above: (A and adi etc.). In fact, they would be shown differently below:

A -> vaHa-
that(demonstrative pronoun)
adi -> vaHa{non-masculine}`
she/it/they
vADu -> vaHa{masculine,singular}`
he
AmeV -> vaHa{fem.,singular}`
she

**3.3 METHOD**

Several levels of analyses and substitution are carried out in anusaaraka. They are at the levels given below:

- morpheme level
- word level
- word group level
- sentence level analysis

For want of space, the processing per se is not discussed any further.

**4. CONCLUSION**

We have discussed the anusaaraka approach to building language access system. It allows rapid development of systems, by separating the analysis based on language and that requiring world knowledge. It takes the view that language encodes information, and the information can be extracted and re-expressed in the target language, by enhancing it with additional notation. It tries to preserve information in the transfer. The user after some training learns to read and understand the text in this "new dialect" of the target language. The output can also be post-edited by a trained user to make it grammatically correct, and stylistically better.

The anusaaraka approach has been successfully used in building systems between five pairs of Indian languages. Work is going on in building an English to Hindi anusaaraka system, which will be a test of building a system between two languages which are far apart.

Anusaarkas follows the principle of substitutibility and reversibility of strings produced, which is nothing but preservation of information while going from a source language to target language. [Because of severe shortage of space, information dynamics has been taken out of this paper. Another attempt will be made to somehow fit the main results of information dynamics with a short introduction, in the final paper, if accepted.]

For narrow subject areas, specialized modules can be built by putting subject domain knowledge into the system, which produce good quality grammatical output. However, it should be remembered, that such modules will work only in narrow areas, and will sometimes go wrong. In such a situation, anusaaraka output will still remain useful.

As pat of future work, work is underway on building an English to Hindi anusaaraka. It will be a further test of the principles and ideas presented here, because they will get applied to two languages which are very different.

## ACKNOWLEDGEMENT

Anusaarakas among Indian languages were built with funding from Ministry of Information Technology, under their program for Technology Development for Indian Languages (TDIL) during 1991-1998. The work was done when the authors were at I.I.T. Kanpur. Currently Satyam Computers Pvt. Ltd. is supporting the authors and the activity for building anusaaraka from English to Hindi. The system so developed will also be available (like the earlier anusaarakas) as "free" open-source software under GPL.

## GLOSSARY

acc. - Accusative marker
erg. - Ergative marker
karaka role - Relation between verb and its arguments
(approximately like theta role)
karta karaka - approx. agent role
TAM - Tense Aspect and Modality
@H - Label to indicate that the ensuing sentence is the anusaaraka output (result of information preserving)
!E - English gloss

## REFERENCES

Natural Language Processing: A Paninian Perspective, Akshar Bharati, Vineet Chaitanya, Rajeev Sangal, Prentice-Hall of India, 1995.

Anusaraka: A Device to Overcome the Language Barrier, V.N. Narayana, Ph.D. thesis, Dept. of CSE, I.I.T. Kanpur, 1994.